\documentclass[letterpaper]{llncs}

\usepackage{times}

\usepackage{index}
\usepackage{setspace} 
\usepackage{hyperref}
\usepackage{graphicx}
\usepackage{hhline}
\usepackage{alltt}
\usepackage{url}
\usepackage{amsmath}
\usepackage{amssymb}
\usepackage{times}
\usepackage{xspace}
\usepackage{color}
\usepackage{theorem}
\usepackage{setspace}
\usepackage{floatflt}
\usepackage{wrapfig}
\usepackage{graphicx}
\usepackage{url}
\usepackage{textcomp}
\usepackage{tikz}
\usetikzlibrary{shapes,backgrounds,arrows,calc,snakes,decorations.text,shapes.geometric,positioning}
\usepackage[normalem]{ulem} 

\usepackage{pdfpages}
\usepackage{url}
\usepackage{mdwlist}
\usepackage{subfigure}
\usepackage{latexsym}
\usepackage{flushend}
\RequirePackage{ifthen}
\RequirePackage{times}      
\RequirePackage{mathptmx}   
\RequirePackage{pifont}     
\usepackage{fancyhdr}
\usepackage{url}

\usepackage{picinpar}

\newcommand{\tbeg}{\langle}
\newcommand{\tend}{\rangle}
\newcommand{\entails}{\models}

\newcommand{\lpnot}{\mbox{not}\;\,}

\newcommand{\hif}{\leftarrow}

\newcommand{\aspindent}{\hspace*{.5in}}

\renewcommand{\epsilon}{\ensuremath{\varepsilon}}

\newcommand{\dom}{\ensuremath{\mathit{Dom}}}
\newcommand{\prob}{\ensuremath{\mathit{Prob}}}
\newcommand{\fs}{\ensuremath{\mathit{Fs}}}
\newcommand{\rs}{\ensuremath{\mathit{Rs}}}
\newcommand{\as}{\ensuremath{\mathit{As}}}
\newcommand{\ps}{\ensuremath{\mathit{Ps}}}
\newcommand{\es}{\ensuremath{\mathit{Es}}}
\newcommand{\arity}{\ensuremath{\mathit{arity}}}
\newcommand{\os}{\ensuremath{\mathit{Os}}}
\newcommand{\init}{\ensuremath{\mathit{Init}}}

\newcommand{\bi}{\begin{itemize}}
\newcommand{\ei}{\end{itemize}}

\def\rrr#1\\{\par
\medskip\hbox{\vbox{\parindent=2em\hsize=6.12in
\hangindent=4em\hangafter=1#1}}}

\usepackage{amsfonts}

\begin{document}

\setcounter{secnumdepth}{2}

\title{PDDL+ Planning via Constraint Answer Set Programming}
\author{Marcello Balduccini\inst{1}
\and
Daniele Magazzeni\inst{2}
\and
Marco Maratea\inst{3}
}
\institute{Drexel University, USA, email: marcello.balduccini@gmail.com
\and
King's College London, UK, email: daniele.magazzeni@kcl.ac.uk
\and
University of Genoa, Italy, email: marco@dibris.unige.it}
\maketitle

\begin{abstract}


PDDL+ is an extension of PDDL that enables modelling planning domains with mixed discrete-continuous dynamics. 
In this paper we present a new approach to PDDL+ planning based on  Constraint Answer Set Programming (CASP), i.e. ASP rules plus numerical constraints. 
To the best of our knowledge, ours is the first attempt to link PDDL+ planning and logic programming. We provide an encoding of PDDL+ models into CASP problems. The encoding can handle non-linear hybrid domains, and represents a solid basis for applying logic programming to PDDL+ planning. 
As a case study, we consider the \textsc{ezcsp} CASP solver and obtain promising results on a set of PDDL+ benchmark problems.

\end{abstract}

\section{Introduction}\label{sec:intro}


Planning in hybrid domains is a challenging problem that
 has found increasing attention in the planning community, mainly motivated by the need to model real-world domains. Indeed, in addition to classical planning, hybrid domains allow for modeling continuous behavior with continuous variables that evolve over time. 
PDDL+ \cite{pddl+} is the extension of the PDDL language that allows for modelling domains with mixed discrete-continuous dynamics, through continuous processes and exogenous events.


\smallskip

 Various techniques and tools have been proposed to deal
 with hybrid domains (as described in Section~\ref{sec:relw}) but few of them are able to handle the full set of PDDL+ features. 
%
%
This motivates the search for new ways to handle PDDL+. To this aim, in this paper we investigate the viability of an approach to PDDL+ planning based on Constraint Answer Set Programming (CASP)~\cite{bbg05}, i.e. set of ASP rules and numerical constraints. We provide an encoding of PDDL+ models into CASP problems, which can handle linear and non-linear domains, and can deal with PDDL+ processes and events. We describe how the different components of a PDDL+ domain can be encoded into CASP. In our encoding, continuous invariants are checked at discretized timepoints, and, following the discretize and validate approach~\cite{upmurphi}, the VAL tool is used to check whether the candidate  solutions are valid or whether granularity needs to be increased. This contribution represents a solid basis for applying logic programming to PDDL+ planning, and opens up the use of CASP solvers for planning in hybrid domains. 

\smallskip
As a case study, we use the CASP solver \textsc{ezcsp}~\cite{bal09}. Experiments performed on PDDL+ benchmarks show that our approach, while not yet competitive with SMTPlan+, outperforms other state-of-the-art PDDL+ planners dReal and UPMurphi.

\smallskip
The paper is structured as follows. We begin with preliminaries on PDDL+ planning and CASP. Next, we present our encoding, and its specialization to the approach of {\sc ezcsp}. These are followed by a discussion of the results of our experiments.
Finally, we discuss related work and draw conclusions.
d


\section{Background} \label{sec:background}
%
Planning 
seeks to select and organize activities in order to achieve
specific goals.
A planner uses a domain model, describing
the actions through their pre- and post-conditions,
and an initial state together with a goal condition. It then
searches for a trajectory through the induced state space,
starting at the initial state and ending in a state satisfying the
goal condition. In richer models, such as hybrid systems, the induced state space can
be given a formal semantics as a timed hybrid automaton,
which means that a plan can synchronise activities between
controlled devices and external events.

 \begin{definition}[Planning Instance \cite{bog14}]
   A planning instance is a pair $I=(\dom,\prob)$, where
   $\dom=(\fs, \rs, \as,$ $\es, \ps, \arity)$ is a tuple consisting of a
   finite set of \emph{function symbols} \fs, a finite set of
   \emph{relation symbols} \rs, a finite set of (durative)
   \emph{actions} \as, a finite set of \emph{events} \es, a finite set
   of \emph{processes} \ps, and a function \arity\ mapping all symbols
   in $\fs\cup\rs$ to their respective arities.
   The triple $\prob=(\os,\init,G)$ consists of a finite set of
   \emph{domain objects} \os, the \emph{initial} state \init, and the
   \emph{goal} specification $G$.
  
 \end{definition}
 %
For
a given planning instance $\Pi$, a \textit{state} of $\Pi$
 consists of a discrete component, described as a set of propositions
 $P$ (the \textit{Boolean fluents}), and a numerical component, described as a set of real
 variables $V$ (the \textit{numerical fluents}). Instantaneous actions are described through
 preconditions (which are conjunctions of propositions in $P$ and/or
 numerical constraints over $V$, and define when an action can be
 applied) and effects (which define how the action modifies the current
 state). \emph{Instantaneous} actions and events are restricted to the
 expression of discrete change. Events have preconditions as for
 actions, but they are used to model exogenous change in the world,
 therefore they are triggered as soon as the preconditions are true. A
 process is responsible for the continuous change of variables, and is
 active as long as its preconditions are true.
 \emph{Durative} actions have three sets of preconditions, representing
 the conditions that must hold when it starts, the invariant that must hold throughout its execution
 and the conditions that must hold at the end of the action. Similarly, a durative action has three sets of
 effects: effects that are applied when the action starts, effects that are applied when the action ends
 and a set of continuous numeric effects which are
 applied continuously while the action is executing. 
\begin{definition}[Plan]
A \textit{plan} for a planning instance $I=((\fs,$ $\rs,$ $\as,$ $\es,$ $\ps,$ $\arity),$ $(\os, \init, G))$ is a finite set of triples $(t,a,d)\in\mathbb{R} \ \times\  \as \ \times\ \mathbb{R}$, where $t$ is a timepoint, $a$ is an action and $d$ is the action duration.
 \end{definition}
Note that processes and events do not appear in a plan, as they are not under the direct control of the planner.






Next, we introduce ASP. Let $\Sigma$ be a signature containing constant, function and
predicate symbols. Terms and atoms are formed as in first-order  logic. A literal is
an atom $a$ or its
classical
negation
$\neg a$. 
A \emph{rule} is a statement of the form:
\begin{equation}\label{eq:rule}
h \hif l_1, \ldots, l_m, \lpnot l_{m+1}, \ldots, \lpnot l_n
\end{equation}
where $h$ and $l_i$'s are literals and $\mbox{\emph{not}}$ is the so-called
\emph{default negation}. The intuitive meaning of the rule is that a
reasoner who believes $\{ l_1, \ldots, l_m \}$ and has no reason to believe
$\{l_{m+1}, \ldots, l_n\}$, has to believe $h$.
We call $h$ the \emph{head} of the rule, and
$\{ l_1, \ldots, l_m, \lpnot l_{m+1},$ $\ldots, \lpnot l_n \}$ the \emph{body}
of the rule.
A rule with an empty body is called a \emph{fact}, and indicates that the head is always true. In that case, the connective $\hif$ is often dropped.
%
A \emph{program} is a set of rules over $\Sigma$.

A set $S$ of literals is \emph{consistent} if no two
complementary literals, $a$ and $\neg a$, belong to $S$. 
A literal $l$ is \emph{satisfied}
by a consistent set of literals $S$ (denoted by $S \entails l$) if $l \in S$. 
If $l$ is not satisfied by $S$,
we write $S \not\entails l$. A set $\{ l_1, \ldots, l_k \}$
of literals is satisfied by a set of literals $S$
($S \entails \{ l_1, \ldots, l_k \}$) if each $l_i$ is satisfied by $S$.

Programs not containing default negation are called \emph{definite}.
A consistent set of literals $S$ is \emph{closed} under a definite
program $\Pi$ if, for every rule of the form (\ref{eq:rule})
such that the body of the rule is satisfied by $S$, the head belongs to $S$.
The \emph{reduct} of an arbitrary program $\Pi$ with respect to a set of
literals $S$, denoted by $\Pi^S$, is the definite program obtained from
$\Pi$ by deleting
every rule $r$ such that $l \in S$ for some expression of
the form $\mathit{not}\ l$ from the body of $r$, and by removing
all expressions $\mathit{not}\ l$ from the bodies of the
remaining rules.

The following definition completes the definition of the semantics of ASP:
\begin{definition}
Consistent set of literals $\mathcal{A}$ is an \emph{answer set} of definite program
$\Pi^*$ if $\mathcal{A}$ is closed under $\Pi^*$ and is set-theoretically
minimal among the sets with that property.
Set $\mathcal{A}$ is an answer set of an arbitrary program
$\Pi$ if it is an answer set of  $\Pi^\mathcal{A}$.
\end{definition}

Variables (identifiers with an uppercase initial) are allowed in ASP programs. A rule containing variables (a \emph{non-ground} rule)
is a shorthand for the set of its \emph{ground instances},
obtained by replacing the variables by all possible ground terms.
Similarly, a non-ground program stands for the set of the ground instances of its rules.

There are other useful shorthands, which we introduce informally to save space. A rule whose head is empty is called \emph{denial}, and states that its body must not be satisfied. A \emph{choice rule} has a head of the form
$\lambda \{ m(\overrightarrow{X}) : \Gamma(\overrightarrow{X}) \} \mu$,
where $\overrightarrow{X}$ is a list of variables, $\lambda$, $\mu$ are non-negative integers, and $\Gamma(\vec{X})$ is a set of literals that may include variables from $\overrightarrow{X}$. A choice rule intuitively states that, in every answer set, the number of literals of the form $m(\overrightarrow{X})$ such that $\Gamma(\overrightarrow{X})$ is satisfied must be between $\lambda$ and $\mu$. If not specified, $\lambda$, $\mu$ default, respectively, to $0$, $\infty$. 
CASP integrates ASP and Constraint Programming (CP) in order to deal with continuous dynamics. In this section  we provide an overview of CP and of its integration in CASP. The central concept of CP is the 
\emph{Constraint Satisfaction Problem (CSP)}, 
which is formally defined as a triple $\tbeg V, D, C \tend$,
 where $V=\{v_1, \ldots, v_n \}$ is a set of variables, $D=\{D_1, \ldots, D_n\}$
 is a set of domains, such that $D_i$ is the domain of variable $v_i$, and $C$
 is a set of constraints. A \emph{solution} to a CSP $\tbeg V, D, C \tend$ is a complete
 assignment (i.e. where a value from the respective domain is assigned to each variable) satisfying every constraint from $C$. For simplicity of presentation, in this paper we allow denoting a CSP\ by its set of constraints, leaving the sets of variables and domains implicitly defined. So, a solution to a set of constraints $C$  is a solution to the CSP implicitly defined by $C$.

There is currently no widely accepted, standardized definition of CASP. 
%
To ensure generality of our results, we introduce  a simplified definition of CASP, defined next, which captures the common traits of the above approaches.
In Section~\ref{sec:impl}, we introduce a specific CASP language to discuss the use case and the experimental results.

\emph{Syntax.} In order to accommodate CP constructs, the language of CASP extends ASP by allowing  \emph{numerical constraints} of the form $x \bowtie y$, where $\bowtie\in\{<,\le,=,\neq, \ge, >\}$, and $x$ and $y$ are \emph{numerical variables}\footnote{Numerical variables are distinct from ASP variables.} or standard mathematical terms possibly containing numerical variables, numerical constants, and ASP variables. Numerical constraints are only allowed in the head of rules.

\emph{Semantics.} Given a numerical constraint $c$, let $\tau(c)$ be a function that maps $c$ to a syntactically legal ASP atom and $\tau^{-1}$ be its inverse\footnote{Technically, $\tau^{-1}$ is a partial inverse whose domain is suitably restricted to ASP atoms that denote a constraint.}. We say that an ASP\ atom $a$ \emph{denotes} a constraint $c$ if $a=\tau(c)$. Function $\tau$ is extended in a natural way to CASP rules and programs. Note that, for every CASP program $\Pi$, $\tau(\Pi)$ is an ASP program. 
Finally, given a set $S$ of ASP literals, let $\gamma(S)$ be the set of ASP\ atoms from $S$ that denote numerical constraints. 
%
The semantics of a CASP program can thus be given by defining the notion of CASP solution, as follows.
\begin{definition}\label{def:casp}
A pair $\tbeg \mathcal{A}, \alpha \tend$ is a \emph{CASP solution} of 
a CASP program $\Pi$ if-and-only-if $\mathcal{A}$ is an answer set of $\tau(\Pi)$ and
$\alpha$ is a solution to $\tau^{-1}(\gamma(\mathcal{A}))$.
\end{definition}

 \section{Encoding PDDL+ Models into CASP Problems}\label{sec:translation}
Our approach to encoding PDDL+ problems in CASP is based on recent research on reasoning about actions and change and action languages.
It builds upon the existing SAT-based
and ASP-based planning approaches and extends them  to hybrid domains.



In reasoning about actions and change, the evolution of a domain over time is often represented by a \emph{transition diagram} (or \emph{transition system}) that represents states and transitions between states through actions. Traditionally, in transition diagrams, actions are instantaneous, and states have no duration and are described by sets of Boolean fluents. Sequences of states characterizing the evolutions of the domain are represented as a sequence of \emph{discrete time steps}, identified by integer numbers,  so that step 0 corresponds to the initial state in the sequence. We extend this view to hybrid domains according to the following principles:
\begin{itemize}
\item
Similarly to PDDL+, a state is characterized by Boolean fluents and numerical fluents.
\item
The flow of actual time is captured by the notion of 
\emph{global time}.
States have a duration, given by the global time at which a state begins and ends. Intuitively, this conveys the intuition that time flows ``within'' the state.
\item
The truth value of Boolean fluents only changes upon state transitions. That is, it is unaffected by the flow of time ``within'' a state. On the other hand, the value of a numerical fluent may change within a state.
\item The global time at which an action occurs is identified with the end time of the state in which the action occurs. 
\item Invariants are checked at the beginning and at the end of every state in which durative actions and processes are in execution. Thus, in order to guarantee soundness we exploit a discretize and validate approach.
\end{itemize}




%
%



Next, we describe the CASP formalization of PDDL+ models. 
We begin by discussing the correspondence between global time and states, and the representation of the values of fluents and of occurrences of actions.

The global time at which the state at step $i$ begins is represented by numerical variable $tstart(i)$. Similarly, the end time is represented by $tend(i)$. 
%
The truth value of Boolean fluent {\sl f} at discrete time step $i$ is represented by literal $holds(f,i)$ if $f$ is true and by $\neg holds(f,i)$ otherwise.
%
For every {\sl numerical fluent n}, we introduce two numerical variables,  representing its value at the beginning and at the end of time step $i$. The variables are $v\_initial(n,i)$ and $v\_final(n,i)$, respectively.
%
The occurrence of an action $a$ at time step $i$ is represented by an atom $occurs(a,i)$.

Additive fluents, whose value is affected by \emph{increase} and \emph{decrease} statements of PDDL+, are represented by introducing numerical variables of the form
$v(contrib(n,s),$ $i)$, 
where $n$ is a numerical fluent, $s$ is a constant denoting a source (e.g., the action that causes the increase or decrease), and $i$ is a time step. The expression denotes the amount of the contribution to fluent $n$ from source $s$ at step $i$. Intuitively, the value of $n$ at the end of step $i$ (encoded by numerical variable $v\_final(n,i)$) is calculated from the values of the individual contributions.
%
%
%
Next, we discuss the encoding of the domain portion of a PDDL+ problem.

\noindent
\textbf{Domain Encoding.}
\emph{(Instantaneous) Actions.} The encoding of the preconditions of actions varies depending on their type. Preconditions on Boolean fluents are encoded by denials. For example, a denial:
$\hif holds(unavail(tk1),I), occurs(re\!fuel\_with(tk1),I)$
states that refuel tank $tk1$ must be available for the corresponding refuel action to occur. (Here and below, ASP variables $I$, $I1$, $I2$ denote time steps.)
Preconditions on numerical fluents are encoded by means of numerical constraints on the corresponding numerical variables. For example, a rule:
$v\_final(height(ball),I)>0 \hif 
occurs(drop(ball),I)$
states that, if $drop(ball)$ is selected to occur, then the height of the ball is required to be greater than $0$ in the preceding state.

The effects of instantaneous actions on Boolean  fluents are captured by rules of the form
$holds(f,I+1) \hif occurs(a,I),$
where $a$ is an action and $f$ is a fluent affected by $a$. The rule states that $f$ is true at the next time step $I+1$ if the action occurs at (the end of) step $I$.
The effects on numerical fluents are represented similarly, but the head of the rule is replaced by a numerical constraint. For example, the rule:
$v\_initial(height(ball),I+1)=10 \hif 
occurs(li\!ft(ball),I)$
states the action of lifting the ball causes its height to be $10$ at the beginning of the state following the occurrence of the action. If the action increases or decreases the value of a numerical fluent, rather than setting it, then a corresponding variable of the form $v(contrib(n,s),i)$ is used in the numerical constraint. The link between contributions and numerical fluent values is established by axioms described later in this section.

\emph{Durative actions.} A durative action $d$ is encoded as two instantaneous actions, $start(d)$ and $end(d)$. The start (end) preconditions of $d$ are mapped to preconditions of $start(d)$ ($end(d)$). The overall conditions are encoded with denials and constraints, as described above in the context of preconditions. Start (end) effects are mapped to effects of $start(d)$ and $end(d)$ actions.
Additionally, $start(d)$ makes fluent $inprogr(d)$ true. The continuous effects of $d$ are made to hold in any state in which $inprogr(d)$ holds. For example, if a $re\!fuel$ action causes the level of fuel in a tank to increase linearly with the flow of time, its effect may be encoded by:
$v(contrib(flevel,re\!fuel),I)=tend(I)-tstart(I) \hif$ 
$holds(inprogr(d),I)$.
The rule intuitively states that, at the end of any state in which $d$ is in progress, the fuel level increases proportionally to the duration of the state. 
The value of the fluent is updated from its set of contributions $S$ by the general constraint,  shown next, which applies to every fluent $F$:
$
v\_final(F,I) = v\_initial(F,I) +\sum_{s\in S} v(contrib(F,s),I). 
$
The fact that the value of numerical fluents stays the same by default throughout the time interval associated with a state is modeled by a rule:
$v\_final(F,I) = v\_initial(F,I) \leftarrow \lpnot ab(F,I),$
which applies to every numerical fluent $F$. Intuitively, $ab(F,I)$ means that $F$ is an exception to the default. 
That is the case when the value of $F$ is being changed by a durative action or process. In those situations, the expression $\lpnot ab(F,I)$ blocks the application of the rule, preventing it from making the final value of $F$ equal to its initial one. Additionally, rules are introduced, which make $ab(F,I)$ true when appropriate. For example, for a durative action $d$ that affects a numerical fluent $f$, the encoding includes a rule:
$ab(f,I) \leftarrow holds(inprogr(d),I).$
In a similar way, the contribution to a numerical fluent by every source is assumed to be $0$ by default. This is guaranteed by the rule:
$v(contrib(F,S),I)=0) \hif \lpnot ab(F,I).$

To keep track of the duration of an action spanning multiple time steps, a rule records the global time at which $d$ begun:
$stime(d)=tend(I) \hif occurs(start(d),I).$
Action $end(d)$ is modeled so that it is automatically triggered after $start(d)$. Finding the time at which the end action occurs, both in terms of time step and global time, is part of the constraint problem to be solved. The rule:
$1\{ occurs(end(d),I2) : I2>I1 \}1 \hif$ 
$occurs(start(d),I1)$
ensures that $end(d)$ will be triggered at some timepoint following $start(d)$. Finally, requirements on the duration of durative actions are  encoded using numerical constraints: if the PDDL+ problem specifies that the duration of $d$ is $\delta$, the requirement is encoded by a rule:
$tend(I)-stime(d)=\delta \hif occurs(end(d),I).$
Intuitively, any CASP solution of the corresponding program will include a specification of when $end(d)$ must occur, both in terms of time step and global time.

It is worth nothing that this encoding extends to multiple occurrences of durative actions in a natural way, by adding, as second argument of instantaneous actions $start(d)$ and $end(d)$, a variable for the timepoint at which the durative action starts (e.g., $start(d,I)$ and $end(d,I)$). Intuitively, this yields multiple, and completely independent, ``copies'' of the durative action, whose effects and termination can be handled accordingly by the encoding presented.

\emph{Processes and Events.}
The encoding of processes and events follows the approach  outlined earlier, respectively, for durative and instantaneous actions. However, their triggering is defined by PDDL+'s \textit{must} semantics, which prescribes that they are triggered as soon as their preconditions are true. In CASP, this is captured by a choice rule combined with numerical constraints. Intuitively, when the Boolean conditions of the process are satisfied, the choice rule states the process will start unless it is inhibited by unsatisfied numerical conditions. Constraints enforced on the numerical conditions capture the latter case. Consider a process corresponding to a falling object, with preconditions $\neg held$ and $height>0$. The choice rule
$1\{ occurs(start(falling),I),$ $is\_false(height>0,I) \}1 \hif$ 
$holds(\neg held,I)$
entails two possible, equally likely, outcomes: the object will either start falling, or be prevented from doing so by the fact that condition $height>0$ is false. The second outcome is possible only if the height is indeed not greater than $0$, 
enforced by:
$v\_final(height,I) \leq 0 \hif is\_false(height>0,I)$.
Generally speaking, given a process with conditions on numerical fluents $n_1,\ldots,n_k$, a choice rule is included in the encoding, with an atom $is\_false(n_{i},I)$ for every $n_i$. A constraint is also added for every condition on some $n_i$. The constraints enforces on $v\_final(n_{i},I)$ the complement of the condition. The treatment of events is similar. 

The encoding is completed by the usual inertia axioms and by rules for preventing gaps between  consecutive states and for handling propagation of fluent values:
%
%
$\{ tstart(I+1) = tend(I).$
$v\_initial(F,I+1) = v\_final(F,I).\}.$

\noindent
\textbf{Problem Encoding.}
The problem portion of the PDDL+ problem is encoded as follows.
\emph{Initial state:}
The encoding of the initial state consists of a set of rules specifying the values of fluents in $P\cup v$ at step $0$.
\emph{Goals:}
The encoding of a goal consists of a set of denials on Boolean fluents and of constraints on numerical fluents, obtained similarly to the encoding of preconditions of actions, discussed earlier.
%
Given a PDDL+ planning instance $I$, by $\Pi(I)$ we denote the CASP encoding of $I$. Next, we turn our attention to the planning task.

\noindent
\textbf{Planning Task.}
Our approach to planning leverages techniques from ASP-based planning.
The planning task is specified by the planning module, $M$, which consists of the single rule:
$
\{ occurs(A,I), occurs(start(D),I) \},
$
where $A$, $D$ are variables ranging over instantaneous actions and durative actions, respectively. The rule intuitively states that any action may occur (or start) at any time step.
It can be shown that the plans for a given maximum time step for a PDDL+ planning instance $I$ are in one-to-one correspondence with the CASP solutions of $\Pi(I) \cup M$. The plan encoded by a CASP solution $A$ can be easily obtained from the atoms of the form $occurs(a,i)$ and from the value assignments to numerical variables $start(i)$ and $end(i)$. Finally, the $\epsilon$-separation\footnote{$\epsilon$-separation requires that interfering actions must be separated by at least a time interval of length $\epsilon$. Hence, two interfering actions $a_1$ and $a_2$ cannot start or end at the same timepoint.
} is handled, as in 
\cite{upmurphi},
by post-processing the plan.
%
It is also worth noting the level of modularity of our approach. In particular, it is straightforward to perform other reasoning tasks besides planning (e.g, a hybrid of planning and diagnostics is often useful for applications) by replacing the planning module by a different one.

%



 \section{Case Study}\label{sec:impl}
\begin{figure}[htb]
\begin{minipage}[c]{0.3\textwidth}

\begin{center}
\includegraphics[clip=true,trim=250 50 190 160,width=\textwidth]{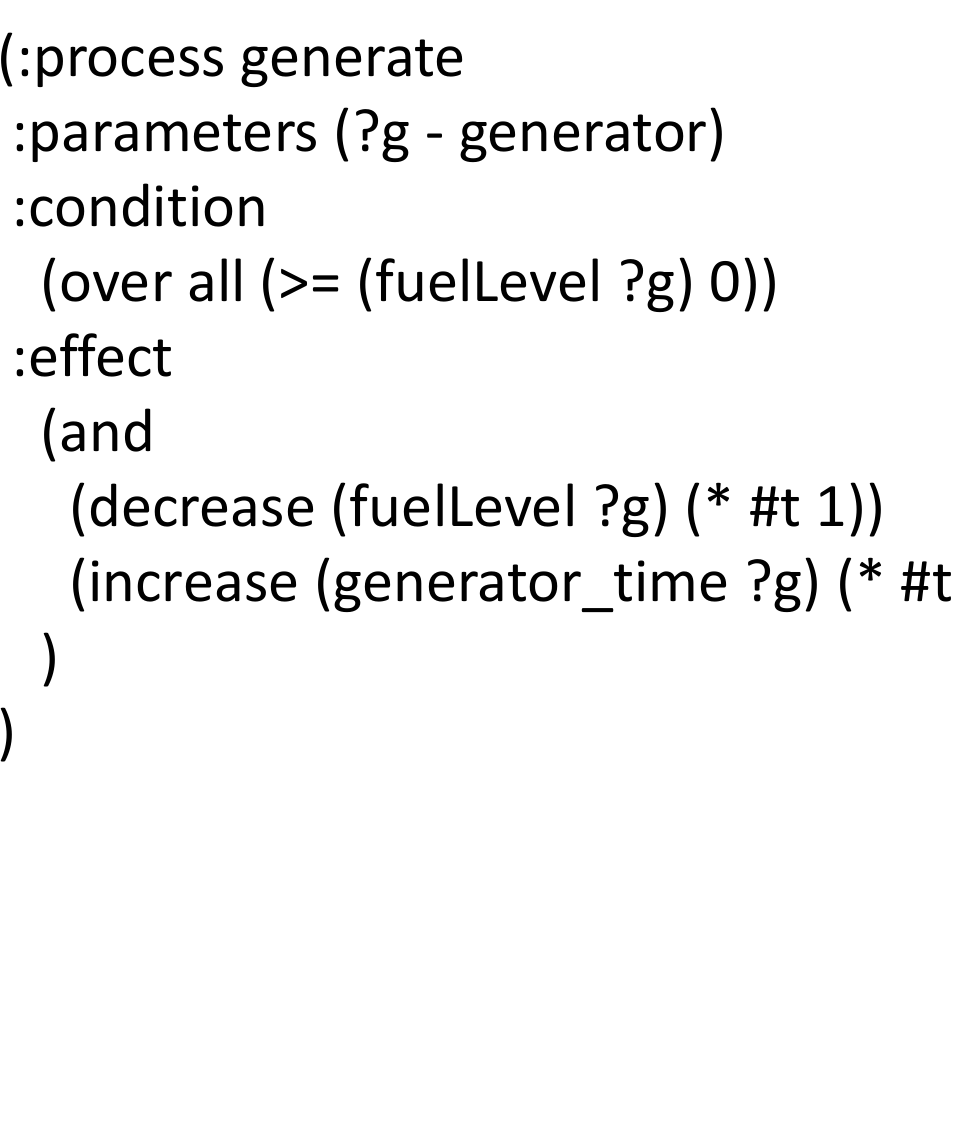}
\end{center}

\end{minipage}
\begin{minipage}[c]{0.7\textwidth}
\begin{center}
\includegraphics[clip=true,trim=0 70 0 60,width=\textwidth]{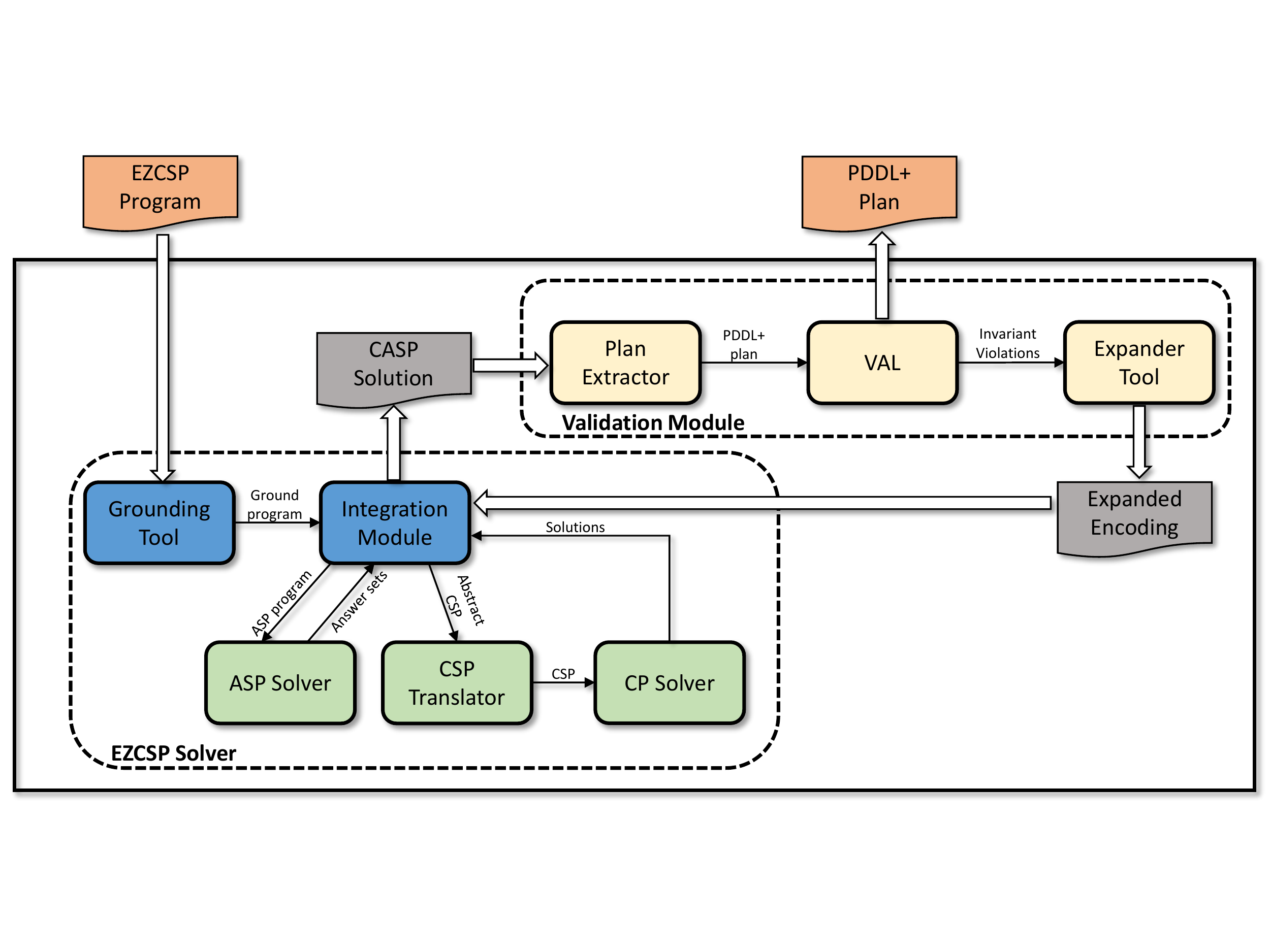}
\end{center}
\end{minipage}
\caption{PDDL+ process from the generator domain \emph{(left)}; Extended solver architecture \emph{(right)}.}\label{fig:domain+arch}
\end{figure}

For our case study,  we have focused on a specific incarnation of CASP, called \textsc{ezcsp} 
\cite{bal09}. 
In \textsc{ezcsp}, numerical constraints are encoded as arguments of the special relation $required$, e.g. $required(start(I+1)=end(I)).$
Encodings of the \textit{generator} 
and \textit{car} domains \cite{bryce} were created as described above, and the architecture of the \textsc{ezcsp} solver was expanded to ensure soundness of the algorithm (see below). 
The complete encodings are omitted due to space considerations. Rather, to illustrate our approach, we present fragments of a possible encoding of process \emph{generate} from the \emph{generator} domain, whose PDDL+ representation is shown in Figure \ref{fig:domain+arch} (left). 
The contribution to the generator's fuel level is encoded by the domain-independent rules discussed earlier,  together with the following problem-specific rules: 
\[
\begin{small}
\begin{array}{l}
cspvar(v(contrib(fuel\_level,decr,generate),I)) \hif step(I). \\
required(v(contrib(fuel\_level,decr,generate),I)>=0) \hif step(I).\\
required(v(contrib(fuel\_level,decr,generate),I)==start(I)-end(I)) \hif\\
\aspindent step(I), 
holds(inprogr(generate),I). \\
\end{array}
\end{small}
\]
From an algorithmic perspective, the \textsc{ezcsp} solver computes CASP solutions of a program $\Pi$ by iteratively (a) using an ASP solver to find an answer set $\mathcal{A}$ of $\Pi$, and (b) using a constraint solver to find the solutions of the CSP encoded by $\mathcal{A}$. 
To account for the discretize and validate approach mentioned earlier, we have extended the \textsc{ezcsp} solver with a validation module, shown in Figure \ref{fig:domain+arch} (right): if step (b) is successful, the tool VAL is called to validate the plan extracted from the CASP solution before returning it. If VAL finds the plan not to be valid, it returns which invariant was violated and at which timepoint. 
If that happens, the \emph{expansion} process occurs, where the encoding is expanded with (i) new numerical variables that represent the value of the involved numerical fluents at that timepoint, and (ii) numerical constraints enforcing the invariant on them.\footnote{Details on the process are omitted to save space.} The CASP solutions for the new encoding are computed again\footnote{Only the solutions of the CSP need to be recomputed.}, and the process is iterated until no invariants are violated.

\section{Experimental Results}\label{sec:exp}


We performed an empirical evaluation of the performance achieved with our approach. The comparison was with the state-of-the-art PDDL+ planners dReal, UPMurphi, and SMTPlan+. 
SpaceEx 
was not considered because it is focused only on plan non-existence.
\begin{table*}[ht!]
\centering
\begin{tabular}{|l|c|c|c|c|c|c|c|c|c|}\hline
\textbf{Domain} & \textbf{Solver}& \textbf{1} & \textbf{2} & \textbf{3} & \textbf{4} & \textbf{5} & \textbf{6} & \textbf{7} & \textbf{8} \\\hline\hline
Gen linear & \textsc{ezcsp}/basic &5.82&2.19&41.77&74.51&114.86&424.80&164.95&--\\\hline
           & \textsc{ezcsp}/heur & 0.28&1.03&4.21&7.25&27.08&43.42&54.83&261.89\\\hline
           & \textsc{ezcsp}/estim & 0.27&0.73&1.64&25.64&77.38&303.75&--&--\\\hline
           & dReal & 3.73&--&--&--&--&--&--&--\\\hline
Gen non-linear & \textsc{ezcsp}/basic & 0.78&3.3&18.27&--&143.19&--&--&*\\\hline
               & \textsc{ezcsp}/heur&0.72&1.62&0.68&1.05&87.95&256.59&238.93&*\\\hline
               & \textsc{ezcsp}/estim&0.81&1.25&0.49&1.19&93.10&50.50&--&*\\\hline
           & dReal & 8.18&--&--&--&--&--&--&--\\\hline
Car linear & \textsc{ezcsp}&0.32&0.31&0.32&0.32&0.32&0.30&0.31&0.31\\\hline
           & dReal & 1.11&1.11&1.15&1.14&1.19&1.13&1.14&1.19\\\hline
Car non-linear & \textsc{ezcsp} &0.71&0.68&0.29&0.39&0.25&0.25&0.26&0.84\\\hline
           & dReal & 58.21&162.60&--&--&--&--&--&--\\\hline
\end{tabular}
\caption{Fixed time step. Results in seconds.
}\label{tab:res}
\end{table*}
\begin{table*}[ht!]
\centering
\begin{tabular}{|l|c|c|c|c|c|c|c|c|c|}\hline
\textbf{Domain} & \textbf{Solver}& \textbf{1} & \textbf{2} & \textbf{3} & \textbf{4} & \textbf{5} & \textbf{6} & \textbf{7} & \textbf{8} \\\hline\hline
Gen linear & \textsc{ezcsp}/basic &1.14&2.71&8.56&12.79&25.90&151.94&96.40&279.81\\\hline
           & \textsc{ezcsp}/heur &0.89&1.92&5.46&9.93&30.79&50.25&67.97&292.22\\\hline
           & \textsc{ezcsp}/estim &0.83&1.55&3.19&26.27&82.32&318.98&--&--\\\hline
           & UPMurphi & 2.02&12.75&91.80&--&--&--&--&--\\\hline
           & SMTPlan+ & 0.06&0.06&0.07&0.09&0.14&0.30&0.93&3.83\\\hline
Gen non-linear & \textsc{ezcsp}/basic &2.30&4.36&42.11&-&152.53&--&--&--\\\hline
               & \textsc{ezcsp}/heur&1.44&2.44&13.10&53.70&88.58&267.11&250.03&--\\\hline
               & \textsc{ezcsp}/estim&0.88&1.89&12.66&54.95&96.47&55.28&--&--\\\hline
           & UPMurphi & --&--&--&--&--&--&--&--\\\hline
           & SMTPlan+ & 0.08&0.08&0.11&0.19&0.37&0.85&2.08&5.22\\\hline
Car linear & \textsc{ezcsp} & 1.01&0.98&1.04&0.99&0.91&0.85&0.88&0.83\\\hline
           & UPMurphi & 0.40&0.38&0.38&0.38&0.41&0.39&0.40&0.41\\\hline
           & SMTPlan+ &0.07&0.06&0.06&0.05&0.05&0.06&0.06&0.06\\\hline
Car non-linear & \textsc{ezcsp} &2.32&1.49&1.14&1.85&1.14&1.18&1.06&2.13\\\hline
           & UPMurphi & 184.88&--&--&--&--&--&--&--\\\hline
           & SMTPlan+ &0.07&0.07&0.07&0.07&0.07&0.07&0.07&0.08\\\hline
\end{tabular}
\caption{Cumulative times. Results in seconds.
}\label{tab:res-cumulative}
\end{table*}
The experimental setup used a VMWare Workstation 12 virtual machine with an single core of a i7-4790K CPU at 4.00GHz and Fedora 22 64 bit. The version of \textsc{ezcsp} was 1.7.4, using
gringo 3.0.5,
clasp 3.1.3,
B-Prolog 7.5,
and GAMS 24.5.7.
B-Prolog was used for all linear problems and GAMS for the non-linear ones. The other systems used were dReal 2.15.11, UPMurphi 3.0.2,
and SMTPlan+ (public version as of Jul 7, 2016).
The experiments were conducted on the \textit{generator} and \textit{car} domains. These are well-known PDDL+ domains and were used as the benchmark set in \cite{bryce}.

The comparison with dReal was based on finding a single plan with a given maximum time step, as discussed in \cite{bryce}. The results are summarized in Table \ref{tab:res}. The comparison with UPMurphi and SMTPlan+ was based on the cumulative times for finding a single plan by progressively increasing the maximum time step. The results are reported in Table \ref{tab:res-cumulative}.
In the tables, entries marked ``-'' indicate a timeout (threshold $600$ sec). Entries marked ``*'' indicate missing entries due to problem size limitations in the free version of GAMS.
It should be noted that none of the instances triggered the expansion process described in the previous section, given that all plans were found to be valid by VAL. Next, we discuss the experimental results obtained for each domain.


 

\textbf{Generator.} 
Our encoding uses Torricelli's law
($v=\sqrt{2gh}$)
to model the transfer of liquid. It should be noted that this is different from the approach used in \cite{bryce}, where a simpler,  but less physically accurate model was used. For a fair comparison with \cite{bryce}, the simpler model was used in reproducing the results for dReal.
The instances were generated by increasing the number of refuel tanks from 1 to 8. The CASP encoding presented earlier is labeled ``\textsc{ezcsp}/basic'' in the table. We also investigated two variants aimed at improving performance of the encoding w.r.t. the treatment of the must semantics. It is not difficult to see that the must semantics may significantly affect performance. 
The encoding labeled to ``\textsc{ezcsp}/heur'' leverages the observation that simple syntactic considerations yield the conclusion that the $generate$ process must start at timepoint $0$. Thus, ``\textsc{ezcsp}/heur'' extends the simpler encoding by a single heuristic stating that action $start(generate)$ must occur immediately. It is interesting to contrast the effects of this domain-specific, encoding-level heuristic with those of the sophisticated, algorithm-level, and yet domain-independent, heuristics used in dReal. 
The encoding labeled ``\textsc{ezcsp}/estim'' takes the observation about the $generate$ process one step further,  replacing the domain-specific heuristic with rules that, in some conditions, can be used to estimate the value of numerical fluents without calling the constraint solver. Compared to dReal and to the previous encoding, the new one is not only encoding-level, but also domain-independent. Furthermore, while dReal's heuristics are specific to the planning task, this approach is task-independent.

The execution times for \textsc{ezcsp} for a fixed maximum time step (Table \ref{tab:res}) were for the most part dominated by ``\textsc{ezcsp}/heur'', which had the best performance in both the linear and the non-linear instances. Remarkably, ``\textsc{ezcsp}/basic'' won over ``\textsc{ezcsp}/estim'' in the linear case, while, as one might have expected, ``\textsc{ezcsp}/estim'' performed better in  the more challenging non-linear case, suggesting that the additional knowledge may be more beneficial in the harder case. The slower times, overall, for the linear case are also somewhat surprising, but are likely due to major differences in the underlying numerical solvers. 
In both the linear and non-linear case, the ``\textsc{ezcsp}/heur'' encoding was substantially faster than dReal, which timed out on all instances except for the first one.

The cumulative times for \textsc{ezcsp} are reported in Table \ref{tab:res-cumulative}. Once again,  ``\textsc{ezcsp}/heur'' had best performance, but ``\textsc{ezcsp}/basic'' had a number of good results in the linear case. Surprisingly,  ``\textsc{ezcsp}/estim'' had the worst performance, timing out on instances $7$ and $8$ in both the linear and non-linear cases. On the other hand, it is interesting to notice that, in the non-linear variant, ``\textsc{ezcp}/estim'' was able to equal and sometimes beat the performance of the other \textsc{ezcsp} encodings before timing out on the last two instances. The reasons for this result are currently unclear and will be the subject of future investigation. 
UPMurphi did not scale as well. In the linear case, only instances $1$-$3$ were solved.
The speedup yielded by \textsc{ezcsp} reached about one order of magnitude before UPMurphi began to time out. In the non-linear case, UPMurphi timed out on all instances. SMTPlan+ outperformed \textsc{ezcsp}, achieving a speedup of about $2$ orders of magnitude and solving one more instance than the latter in the non-linear case.

\textbf{Car.} 
The instances were obtained by progressively increasing the range of allowed accelerations (velocities in the linear version) from $[-1,1]$ to $[-8,8]$. The CASP encoding leveraged no heuristics. 
%
As shown in Table \ref{tab:res}, the execution times for \textsc{ezcsp} were
about $3$ times faster than dReal in the linear case and orders of magnitude better in the non-linear case, where dReal timed out in instances 3-8. The run-times of \textsc{ezcsp} showed no significant growth in either case.
The comparison with UPMurphi on cumulative times shows some interesting behavior. In the linear case, \textsc{ezcsp} was, in fact, about $2$-$3$ times slower than UPMurphi. 
On the other hand, \textsc{ezcsp} outperformed UPMurphi in the non-linear case, 
where
UPMurphi only solved the first instance with a time 
nearly $2$ orders of magnitude slower than \textsc{ezcsp}. SMTPlan+ outperformed \textsc{ezcsp} in this domain as well, with speedups of a little over $1$ order of magnitude.

We believe the empirical results demonstrate that our approach is promising, beating by a good margin the state-of-the-art planners with the exception of SMTPlan+. As for the difference in performance with SMTPlan+, a thorough algorithmic and architectural comparison has yet to be conducted due to SMTPlan+ being fairly recent. At this point, various explanations are possible.  First of all, the \textsc{ezcsp} encoding was designed for clarity and elaboration tolerance rather than speed. Next, our approach currently lacks important optimizations present in SMTPlan+ -- e.g., incremental grounding and multi-threading. Lastly, the specific numerical solvers used may also play an important role.   



\section{Related Work}\label{sec:relw}

To the best of our knowledge, ours is the first attempt to link PDDL+ planning and logic programming. Various techniques and tools have been proposed to deal
with hybrid domains using other techniques.
Most of them are unable to handle the full set of PDDL+ features, namely non-linear domains with processes and events. 
For instance, dReach \cite{bryce} leverages SMT to plan in hybrid systems, but does not provide a direct translation from PDDL+ and does not handle exogenous events.

SMTPlan+ \cite{smtplan+} is another approach based on a translation to SMT, but it supports the full PDDL+ language, and featured excellent performance in our evaluation.

From the model checking and control communities, a number of approaches have been proposed (e.g., \cite{bog14}). 
Similarly,  \textit{falsification} of hybrid systems tries to guide the search towards the error states, which can be easily cast as a planning problem.
However, while all of these works aim to address hybrid systems, they do not handle PDDL+ models. 

UPMurphi~\cite{upmurphi} is capable of handling the full set of PDDL+ features, although it is very limited in scalability as it performs blind search. The approach proposed in this paper is similar to TM-LPSAT~\cite{TM-LPSAT}. However, TM-LPSAT assumes linear continuous change, while we tackle problems with nonlinear dynamics.

For what concerns CASP solvers, a high level view of the languages and solving techniques can be found in \cite{Lierler14}. Of the available systems, \textsc{ezcsp} is, to the best of our knowledge, the only one supporting both non-linear constraints, required for modeling non-linear continuous change, and real numbers. 
 \section{Conclusions}\label{sec:concl}

We have presented a novel approach to PDDL+ planning based on CASP languages, providing a solid basis for applying logic programming to PDDL+ planning. Experiments on 
well-known domains, some involving non-linear continuous change, showed that
our approach outperforms most comparable state-of-the-art PDDL+ planners.
Basing our approach on CASP also makes it amenable to be expanded to handle uncertainty about the initial situation or the effects of actions, and to reasoning tasks other than planning. In the future, we plan to investigate these aspects, and to conduct a thorough algorithmic and architectural comparison with SMTPlan+.





\bibliographystyle{splncs03}
\bibliography{101-biblio-mb,102-biblio-dm,103-biblio-mm}
\end{document}